\documentclass[10pt,twocolumn,letterpaper]{article}

\usepackage{cvpr}
\usepackage{times}
\usepackage{epsfig}
\usepackage{graphicx}
\usepackage{amsmath}
\usepackage{amssymb}

\usepackage{array}
\usepackage{tabulary}
\usepackage{algorithm,algorithmicx,algpseudocode}
\usepackage{mathtools, nccmath}
\usepackage{booktabs}
\usepackage{comment}
\usepackage{bbm}
\usepackage{footnote}
\usepackage{subfig}
\usepackage{multirow}

\usepackage[font=small]{caption}
\usepackage{tablefootnote}

\usepackage[breaklinks=true,bookmarks=false]{hyperref}

\cvprfinalcopy 

\def\cvprPaperID{5104} 

\ifcvprfinal\fi
\begin{document}

\title{Reinforced Cross-Modal Matching and Self-Supervised Imitation Learning 
\\ for Vision-Language Navigation}

\author{Xin Wang$^1$ \quad Qiuyuan Huang$^2$ \quad Asli Celikyilmaz$^2$ \quad Jianfeng Gao$^2$ \quad Dinghan Shen$^3$ \\
Yuan-Fang Wang$^1$ \quad William Yang Wang$^1$ \quad Lei Zhang$^2$\\
$^1$University of California, Santa Barbara \quad $^2$Microsoft Research, Redmond \quad $^3$Duke University \\ 
{\tt\small \{xwang,yfwang,william\}@cs.ucsb.edu} \\
{\tt\small \{qihua,aslicel,jfgao,leizhang\}.microsoft.com, dinghan.shen@duke.edu}
}
\maketitle

\begin{abstract}
Vision-language navigation (VLN) is the task of navigating an embodied agent to carry out natural language instructions inside real 3D environments. In this paper, we study how to address three critical challenges for this task: the cross-modal grounding, the ill-posed feedback, and the generalization problems. 
First, we propose a novel Reinforced Cross-Modal Matching (RCM) approach that enforces cross-modal grounding both locally and globally via reinforcement learning (RL). Particularly,  a matching critic is used to provide an intrinsic reward to encourage global matching between instructions and trajectories, and a reasoning navigator is employed to perform cross-modal grounding in the local visual scene. 
Evaluation on a VLN benchmark dataset shows that our RCM model significantly outperforms previous methods by 10\% on SPL and achieves the new state-of-the-art performance. 
To improve the generalizability of the learned policy, we further introduce a Self-Supervised Imitation Learning (SIL) method to explore unseen environments by imitating its own past, good decisions. We demonstrate that SIL can approximate a better and more efficient policy, which tremendously minimizes the success rate performance gap between seen and unseen environments (from 30.7\% to 11.7\%).
\end{abstract}

\section{Introduction}
Recently, vision-language grounded embodied agents have received increased attention~\cite{minos,kolve2017ai2,embodiedqa} due to their popularity in many intriguing real-world applications, \eg, in-home robots and personal assistants.
Meanwhile, such an agent pushes forward visual and language grounding by putting itself in an active learning scenario through first-person vision.
In particular, Vision-Language Navigation (VLN)~\cite{anderson2018vision} is the task of navigating an agent inside real environments by following natural language instructions.
VLN requires a deep understanding of linguistic semantics,  visual perception, and most importantly, the alignment of the two. The agent must reason about the vision-language dynamics in order to move towards the target that is inferred from the instructions.

\begin{figure}
\begin{center}
\includegraphics[width=0.5\textwidth]{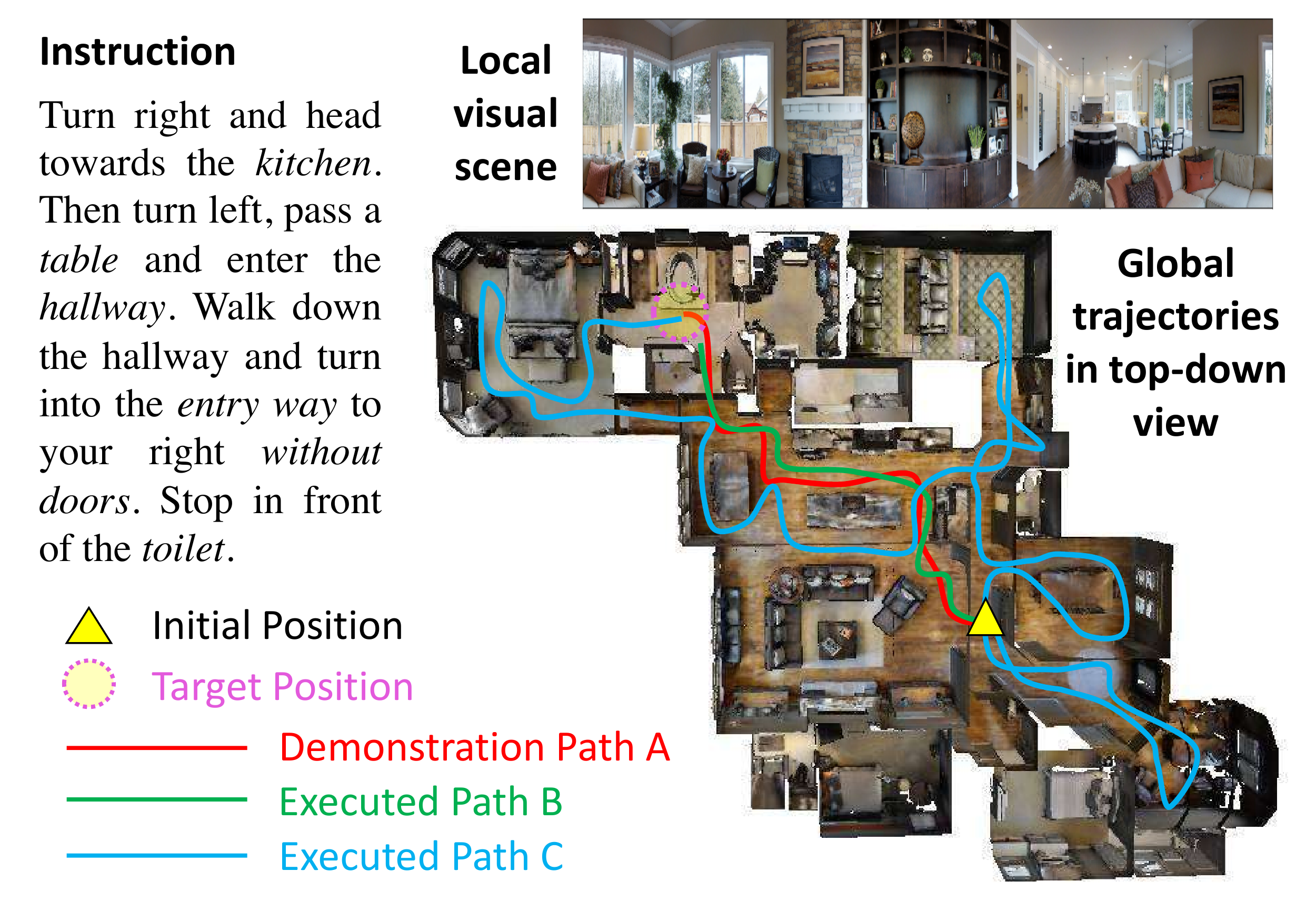}  
\end{center}
\vspace{-2ex}
\caption{Demonstration of the VLN task. The instruction, the local visual scene, and the global trajectories in a top-down view is shown. The agent does not have access to the top-down view. Path A is the demonstration path following the instruction. Path B and C are two different paths executed by the agent.}
\label{fig:intro}
\end{figure}

VLN presents some unique challenges.  
First, reasoning over visual images and natural language instructions can be difficult. 
As we demonstrate in Figure~\ref{fig:intro}, to reach a destination, the agent needs to ground an instruction in the local visual scene, represented as a sequence of words, as well as match the instruction to the visual trajectory in the global temporal space. 
Secondly, except for strictly following expert demonstrations, the feedback is rather coarse, since the ``Success" feedback is provided only when the agent reaches a target position, completely ignoring whether the agent has followed the instructions (\eg, Path A in Figure~\ref{fig:intro}) or followed a random path to reach the destination (\eg, Path C in Figure~\ref{fig:intro}). Even a ``good" trajectory that matches an instruction can be considered unsuccessful if the agent stops marginally earlier than it should be (\eg, Path B in Figure~\ref{fig:intro}). 
An ill-posed feedback can potentially deviate from the optimal policy learning. 
Thirdly, existing work suffers from the generalization problem, causing a huge performance gap between seen and unseen environments. 

In this paper, we propose to combine the power of reinforcement learning (RL) and imitation learning (IL) to address the challenges above. First, we introduce a novel Reinforced Cross-Modal Matching (RCM) approach that enforces cross-modal grounding both locally and globally via RL. 
Specifically, we design a \textit{reasoning navigator} that learns the cross-modal grounding in both the textual instruction and the local visual scene, so that the agent can infer which sub-instruction to focus on and where to look at. 
From the global perspective, we equip the agent with a \textit{matching critic} that evaluates an executed path by the probability of reconstructing the original instruction from it, which we refer to as the cycle-reconstruction reward. Locally, the cycle-reconstruction reward provides a fine-grained intrinsic reward signal to encourage the agent to better understand the language input and penalize the trajectories that do not match the instructions. For instance, using the proposed reward, Path B is considered better than Path C (see Figure~\ref{fig:intro}). 

Being trained with the intrinsic reward from the matching critic and the extrinsic reward from the environment, the reasoning navigator learns to ground the natural language instruction on both local spatial visual scene and global temporal visual trajectory. Our RCM model significantly outperforms the existing methods and achieves new state-of-the-art performance on the Room-to-Room (R2R) dataset. 

Our experimental results indicate a large performance gap between seen and unseen environments.  
To narrow the gap, we propose an effective solution to explore unseen environments with self-supervision. This technique is valuable because it can facilitate lifelong learning and adaption to new environments. For example, domestic robots can explore a new home it arrives at and iteratively improve the navigation policy by learning from previous experience. Motivated by this fact, we introduce a Self-Supervised Imitation Learning (SIL) method in favor of exploration on unseen environments that do not have labeled data. The agent learns to imitate its own past, good experience. Specifically, in our framework, the navigator performs multiple roll-outs, of which good trajectories (determined by the matching critic) are stored in the replay buffer and later used for the navigator to imitate. In this way, the navigator can approximate its best behavior that leads to a better policy. 
To summarize, our contributions are mainly three-fold:
\begin{itemize}
    \item We propose a novel Reinforced Cross-Modal Matching (RCM) framework that utilizes both extrinsic and intrinsic rewards for reinforcement learning, of which we introduce a cycle-reconstruction reward as the intrinsic reward to enforce the global matching between the language instruction and the agent's trajectory.
    \item Experiments show that RCM achieves the new state-of-the-art performance on the R2R dataset, and among the prior art, is ranked first\footnote{As of November 16th, 2018.} in the VLN Challenge in terms of SPL, the most reliable metric for the task.
    \item We introduce a new evaluation setting for VLN,  where exploring unseen environments prior to testing is allowed, 
    and then propose a Self-Supervised Imitation Learning (SIL) method for exploration with self-supervision, whose effectiveness and efficiency are validated on the R2R dataset.
\end{itemize}


\section{Related Work}
\paragraph{Vision-and-Language Grounding}
Recently, researchers in both computer vision and natural language processing are striving to bridge vision and natural language towards a deeper understanding of the world~\cite{xu2015show,wang2018AREL,karpathy2015deep,chen2015mind,hu2016segmentation,tapaswi2016movieqa,huang2018turbo}, \eg, captioning an image or a video with natural language~\cite{lrcn2014,fang2015captions,vinyals2015show,wang2018video,yang2016stacked,yu2016video,wang2018watch} or localizing desired objects within an image given a natural language description~\cite{plummer2015flickr30k,hu2016natural,yu2016modeling,zhang2018man}. Moreover, visual question answering~\cite{VQA} and visual dialog~\cite{visdial} aim to generate one-turn or multi-turn response by grounding it on both visual and textual modalities. 
However, those tasks focus on passive visual perception in the sense that the visual inputs are usually fixed. In this work, we are particularly interested in solving the dynamic multi-modal grounding problem in both temporal and spatial spaces. Thus, we focus on the task of vision-language navigation (VLN)~\cite{anderson2018vision} which requires the agent to actively interact with the environment.

\paragraph{Embodied Navigation Agent}
Navigation in 3D environments~\cite{zhu2017target,mirowski2016learning,mousavian2018visual,hemachandra2015learning} is an essential capability of a  mobile intelligent system that functions in the physical world. In the past two years, a plethora of tasks and evaluation protocols~\cite{minos,kolve2017ai2,song2016ssc,xiazamirhe2018gibsonenv,anderson2018vision} have been proposed as summarized in \cite{anderson2018evaluation}. VLN~\cite{anderson2018vision} focuses on language-grounded navigation in the real 3D environment. 
In order to solve the VLN task, Anderson \etal~\cite{anderson2018vision} set up an attention-based sequence-to-sequence baseline model. Then Wang \etal~\cite{wang2018look} introduced a hybrid approach that combines model-free and model-based reinforcement learning (RL) to improve the model's generalizability. Lately, Fried \etal~\cite{fried2018speaker} proposed a speaker-follower model that adopts data augmentation, panoramic action space and modified beam search for VLN, establishing the current state-of-the-art performance on the Room-to-Room dataset. Extending prior work, we propose a Reinforced Cross-Modal Matching (RCM) approach to VLN. The RCM model is built upon \cite{fried2018speaker} but differs in many significant aspects: (1) we combine a novel multi-reward RL with imitation learning for VLN while Speaker-Follower models~\cite{fried2018speaker} only uses supervised learning as in \cite{anderson2018vision}. (2) Our reasoning navigator performs cross-modal grounding rather than the temporal attention mechanism on single-modality input.
(3) Our matching critic is similar to Speaker in terms of the architecture design, but the former is used to
provide the cycle-reconstruction intrinsic reward for both RL and SIL training while the latter is used to augment training data for supervised learning. Moreover, we introduce a self-supervised imitation learning method for exploration in order to explicitly address the generalization issue, which is a problem not well-studied in prior work.
Concurrent to our work, \cite{thomason2018shifting,ke2017tactile,ma2019self,ma2019regretful} studies the VLN tasks from various aspects, and \cite{nguyen2018vision} introduces a variant of the VLN task to find objects by requesting language assistance when needed. Note that we are the first to propose to explore unseen environments for the VLN task.

\paragraph{Exploration}
Much work has been done on improving exploration~\cite{bellemare2016unifying,gaosurvey,houthooft2016vime,ostrovski2017count,tang2017exploration}
because the trade-off between exploration and exploitation is one of the fundamental challenges in RL. The agent needs to exploit what it has learned to maximize reward and explore new territories for better policy search.
Curiosity or uncertainty has been used as a signal for exploration~\cite{schmidhuber1991adaptive,strehl2008analysis,lipton2016efficient,pathak2017curiosity}. Most recently, Oh \etal~\cite{oh2018self} proposed to exploit past good experience for better exploration in RL and theoretically justified its effectiveness.
Our Self-Supervised Imitation Learning (SIL) method shares the same spirit. But instead of testing on games, we adapt SIL and validate its effectiveness and efficiency on the more practical task of VLN. 

\section{Reinforced Cross-Modal Matching}
\subsection{Overview}
\begin{figure}
\begin{center}
\includegraphics[width=0.42\textwidth]{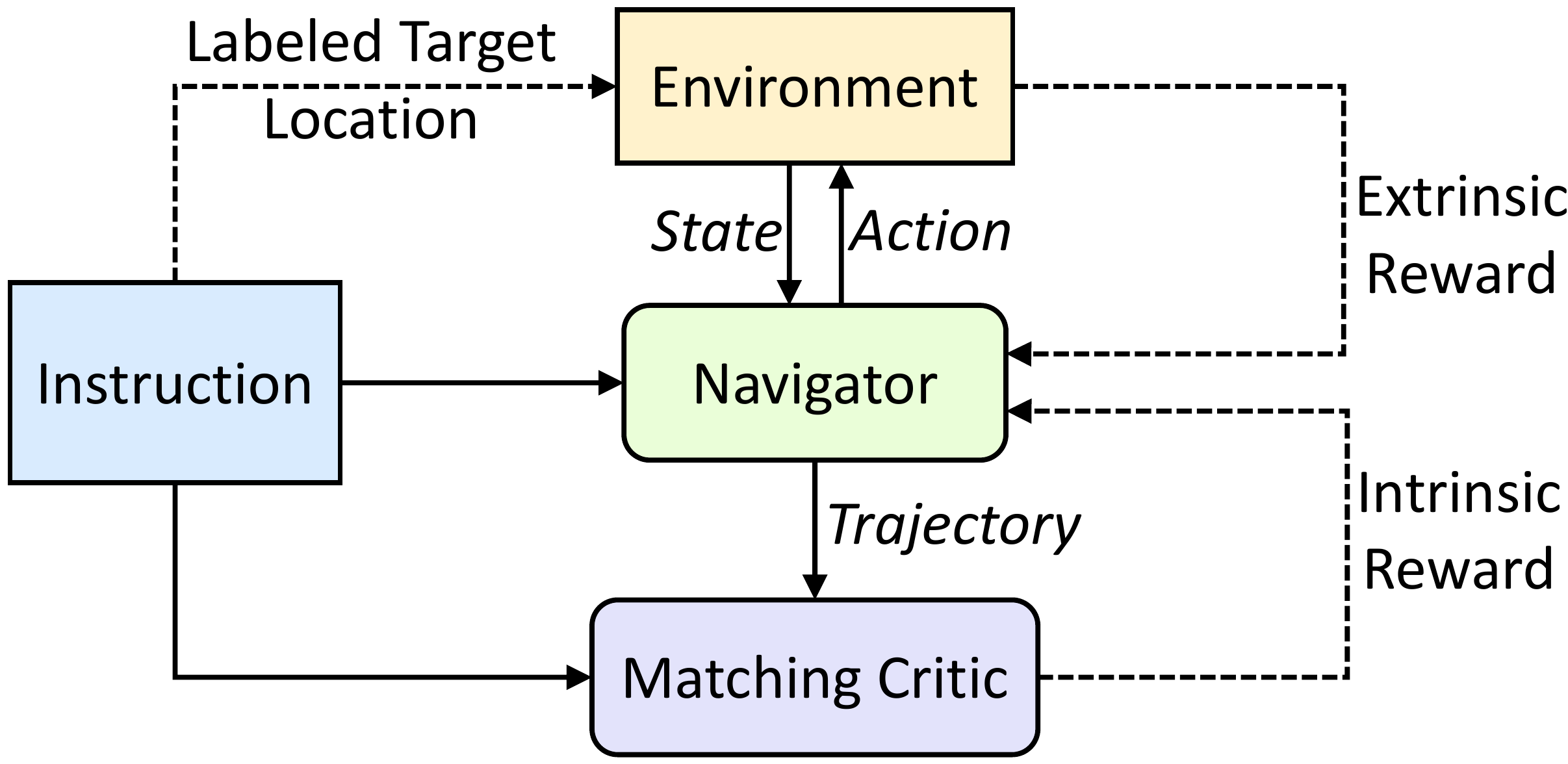}  
\end{center}
\caption{Overview of our RCM framework.}
\label{fig:overview}
\end{figure}

Here we consider an embodied agent that learns to navigate inside real indoor environments by following natural language instructions. 
The RCM framework mainly consists of two modules (see Figure~\ref{fig:overview}): a \textbf{reasoning navigator} $\pi_{\theta}$ and a \textbf{matching critic} $V_{\beta}$.
Given the initial state $s_0$ and the natural language instruction (a sequence of words) $\mathcal{X} = {x_1, x_2, ..., x_n}$, the reasoning navigator learns to perform a sequence of actions ${a_1, a_2, ..., a_T} \in \mathcal{A}$, which generates a trajectory $\tau$, in order to arrive at the target location $s_{target}$ indicated by the instruction $\mathcal{X}$. 
The navigator interacts with the environment and perceives new visual states as it executes actions. To promote the generalizability and reinforce the policy learning, we introduce two reward functions: an \textbf{extrinsic reward} that is provided by the environment and measures the success signal and the navigation error of each action, and an \textbf{intrinsic reward} that comes from our matching critic and measures the alignment between the language instruction $\mathcal{X}$ and the navigator's trajectory $\tau$. 

\subsection{Model}\label{sec:model}
Here we discuss the reasoning navigator and matching critic in details, both of which are end-to-end trainable. 

\subsubsection{Cross-Modal Reasoning Navigator} 
 
The navigator $\pi_{\theta}$ is a policy-based agent that maps the input instruction $\mathcal{X}$ onto a sequence of actions $\{a_t\}_{t=1}^T$. At each time step $t$, the navigator receives a state $s_t$ from the environment and needs to ground the textual instruction in the local visual scene. Thus, we design a cross-modal reasoning navigator that learns the trajectory history, the focus of the textual instruction, and the local visual attention in order, which forms a cross-modal reasoning path to encourage the local dynamics of both modalities at step $t$.  

\begin{figure}
\begin{center}
\includegraphics[width=0.5\textwidth]{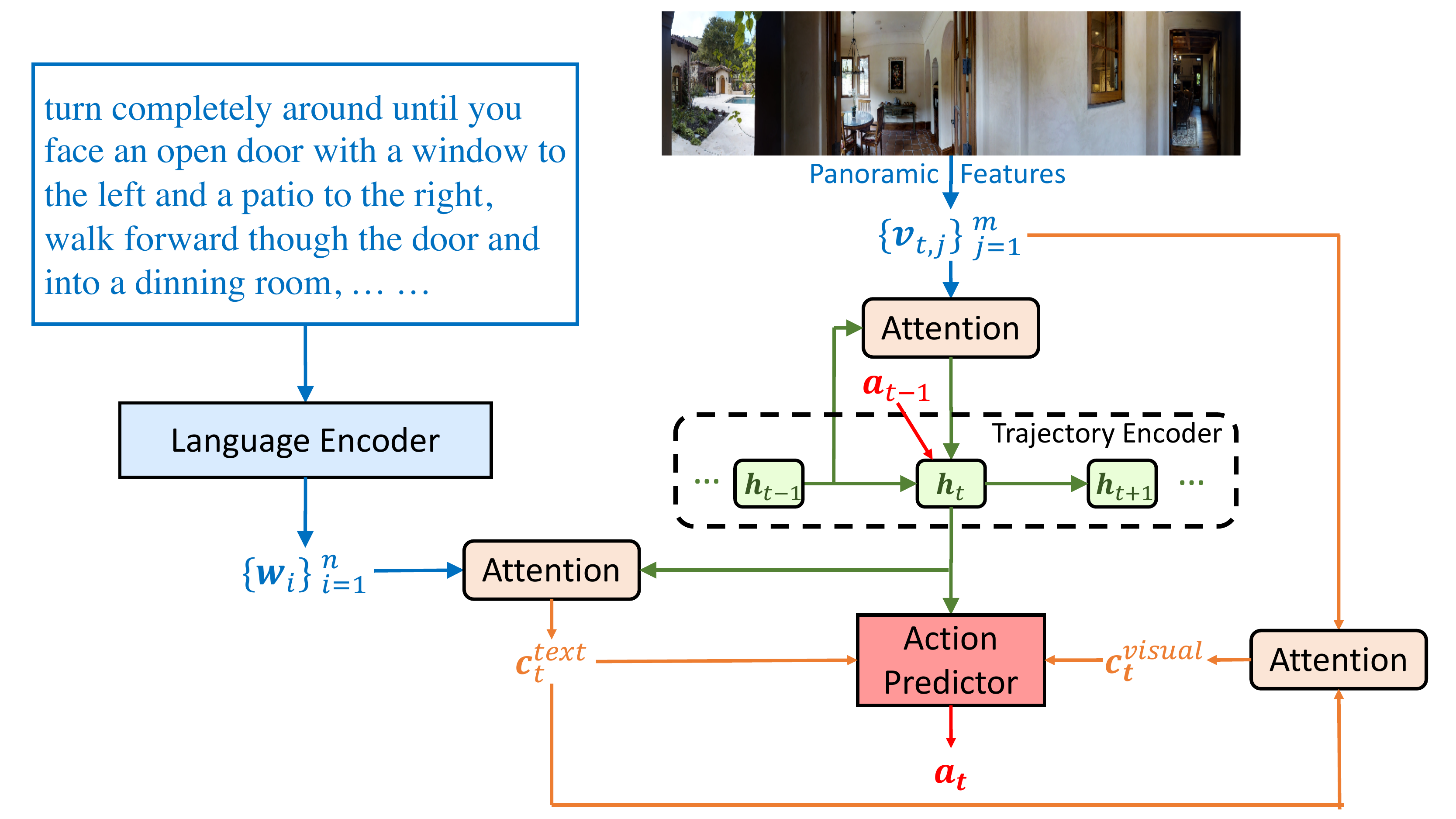}  
\end{center}
\vspace{-2ex}
\caption{Cross-modal reasoning navigator at step $t$.}
\label{fig:nav}
\vspace{-1ex}
\end{figure}

Figure~\ref{fig:nav} shows the unrolled version of the navigator at time step $t$. Similar to \cite{fried2018speaker}, we equip the navigator with a panoramic view, which is split into image patches of $m$ different viewpoints, so the panoramic features that are extracted from the visual state $s_t$ can be represented as $\{v_{t,j}\}_{j=1}^{m}$, where $v_{t,j}$ denotes the pre-trained CNN feature of the image patch at viewpoint $j$. 

\paragraph{History Context} Once the navigator runs one step, the visual scene would change accordingly. The history of the trajectory $\tau_{1:t}$ till step $t$ is encoded as a history context vector $h_t$ by an attention-based trajectory encoder LSTM~\cite{hochreiter1997long}:
\begin{equation}
    h_t = LSTM([v_t, a_{t-1}], h_{t-1})
\end{equation}
where $a_{t-1}$ is the action taken at previous step, and $v_t = \sum_j \alpha_{t,j} v_{t, j}$, the weighted sum of the panoramic features. $\alpha_{t,j}$ is the attention weight of the visual feature $v_{t, j}$, representing its importance with respect to the previous history context $h_{t-1}$. Note that we adopt the dot-product attention~\cite{vaswani2017attention} hereafter, which we denote as (taking the attention over visual features above for an example) 
\begin{align}
    v_t &= attention(h_{t-1}, \{v_{t,j}\}_{j=1}^m) \\
    &= \sum_j softmax(h_{t-1} W_h (v_{t,j} W_v)^T) v_{t,j}
\end{align}
where $W_h$ and $W_v$ are learnable projection matrices.

\paragraph{Visually Conditioned Textual Context} Memorizing the past can enable the recognition of the current status and thus understanding which words or sub-instructions to focus on next. Hence, we further learn the textual context $c_t^{text}$ conditioned on the history context $h_t$. We let a language encoder LSTM to encode the language instruction $\mathcal{X}$ into a set of textual features $\{w_i\}_{i=1}^n$. Then at every time step, the textual context is computed as 
\begin{equation}
    c_t^{text} = attention(h_t, \{w_i\}_{i=1}^n)
\end{equation}
Note that $c_t^{text}$ weighs more on the words that are more relevant to the trajectory history and the current visual state.

\paragraph{Textually Conditioned Visual Context} Knowing where to look at requires a dynamic understanding of the language instruction; so we compute the visual context $c_t^{visual}$ based on the textual context $c_t^{text}$:
\begin{equation}
    c_t^{visual} = attention(c_t^{text}, \{v_j\}_{j=1}^m)
\end{equation}

\paragraph{Action Prediction}
In the end, our action predictor considers the history context $h_t$, the textual context $c_t^{text}$, and the visual context $c_t^{visual}$, and decides which direction to go next based on them. It calculates the
probability $p_k$ of each navigable direction using a bilinear dot product as follows:
\begin{equation}
    p_k = softmax([h_t, c_t^{text}, c_t^{visual}]W_c (u_k W_u)^T)
\end{equation}
where $u_k$ is the action embedding that represents the $k$-th navigable direction, which is obtained by concatenating an appearance feature vector (CNN feature vector extracted from the image patch around that view angle or direction) and a 4-dimensional orientation feature vector $[sin \psi; cos \psi; sin \omega; cos \omega]$, where $\psi$ and $\omega$ are the heading and elevation angles respectively. 
The learning objectives for training the navigator are introduced in Section~\ref{sec:learning}.

\begin{figure}
\begin{center}
\includegraphics[width=0.45\textwidth]{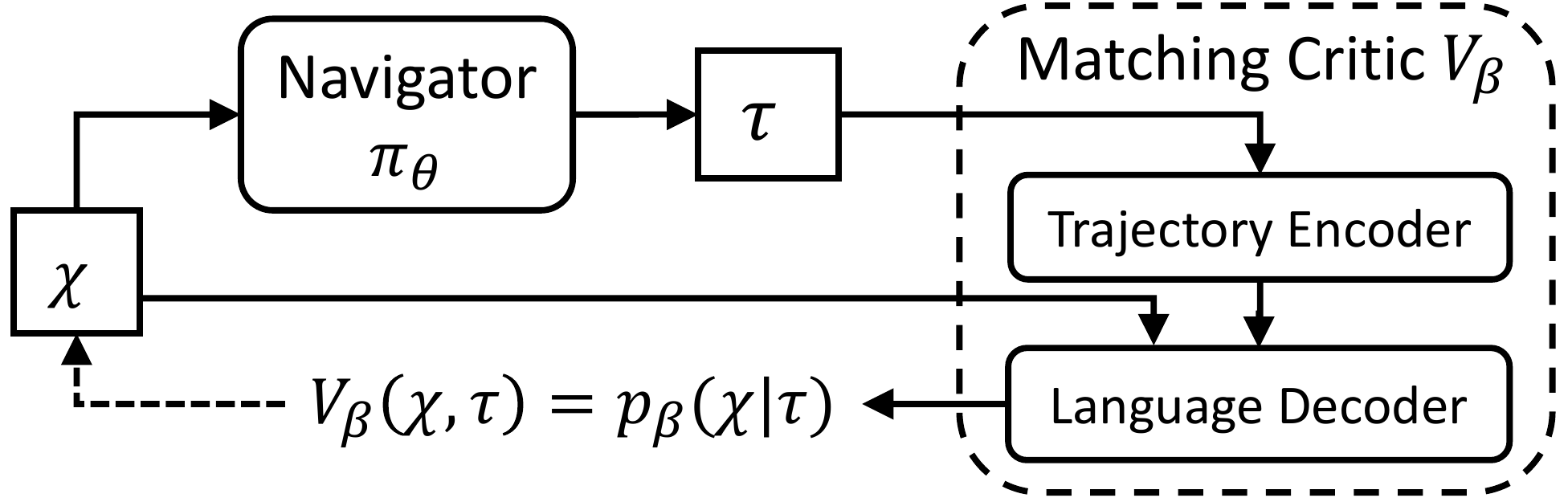}  
\end{center}
\caption{Cross-modal matching critic that provides the cycle-reconstruction intrinsic reward.}
\label{fig:critic}
\end{figure}

\subsubsection{Cross-Modal Matching Critic}\label{sec:critic}
In addition to the extrinsic reward signal from the environment, we also derive an intrinsic reward $R_{intr}$ provided by the matching critic $V_{\beta}$ to encourage the global matching between the language instruction $\mathcal{X}$ and the navigator $\pi_{\theta}$'s trajectory $\tau = \{<s_1, a_1>, <s_2, a_2>, ..., <s_T, a_T>\}$:
\begin{equation}
    R_{intr} = V_{\beta}(\mathcal{X}, \tau) = V_{\beta}(\mathcal{X}, \pi_{\theta}(\mathcal{X})) 
\end{equation}
One way to realize this goal is to measure the cycle-reconstruction reward $p(\hat{\mathcal{X}} = \mathcal{X} | \pi_{\theta}(\mathcal{X}))$, the probability of reconstructing the language instruction $\mathcal{X}$ given the trajectory $\tau = \pi_{\theta}(\mathcal{X})$ executed by the navigator. The higher the probability is, the better the produced trajectory is aligned with the instruction. 

Therefore as shown in Figure~\ref{fig:critic}, we adopt an attention-based sequence-to-sequence language model as our matching critic $V_{\beta}$, which encodes the trajectory $\tau$ with a trajectory encoder and produces the probability distributions of generating each word of the instruction $\mathcal{X}$ with a language decoder. Hence the intrinsic reward
\begin{equation}
\label{eq:intrinsic}
    R_{intr} = p_{\beta}(\mathcal{X} | \pi_{\theta}(\mathcal{X})) = p_{\beta}(\mathcal{X} | \tau) 
\end{equation}
which is normalized by the instruction length $n$.
In our experiments, the matching critic is pre-trained with human demonstrations (the ground-truth instruction-trajectory pairs $<\mathcal{X}^*, \tau^*>$) via supervised learning. 

\subsection{Learning}\label{sec:learning}
In order to quickly approximate a relatively good policy, we use the demonstration actions to conduct supervised learning with maximum likelihood estimation (MLE). The training loss $L_{sl}$ is defined as 
\begin{equation}
    L_{sl} = - \mathbb{E}[\log(\pi_{\theta}(a_t^* | s_t))]
\end{equation}
where $a_t^*$ is the demonstration action provided by the simulator. Warm starting the agent with supervised learning can ensure a relatively good policy on the seen environments. But it also limits the agent's generalizability to recover from erroneous actions in unseen environments, since it only clones the behaviors of expert demonstrations.

To learn a better and more generalizable policy, we then switch to reinforcement learning and introduce the extrinsic and intrinsic reward functions to refine the policy from different perspectives. 

\paragraph{Extrinsic Reward} 
A common practice in RL is to directly optimize the evaluation metrics. Since the objective of the VLN task is to successfully reach the target location $s_{target}$, we consider two metrics for the reward design. 
The first metric is the relative navigation distance similar to \cite{wang2018look}. We denote the distance between $s_t$ and $s_{target}$ as $\mathcal{D}_{target}(s_t)$. Then the immediate reward $r(s_t,a_t)$ after taking action $a_t$ at state $s_t$ ($t < T$) becomes:
\begin{equation}
    r(s_t,a_t) = \mathcal{D}_{target}(s_{t}) - \mathcal{D}_{target}(s_{t+1}), \quad t < T
\end{equation}
This indicates the reduced distance to the target location after taking action $a_t$. 
Our second choice considers the ``Success" as an additional criterion. If the agent reaches a point within a threshold measured by the distance $d$ from the target ($d$ is preset as 3m in the R2R dataset), then it is counted as ``Success". Particularly, the immediate reward function at last step $T$ is defined as 
\begin{equation}
    r(s_T,a_T) = \mathbbm{1}(\mathcal{D}_{target}(s_{T}) \leq d)
\end{equation}
where $\mathbbm{1}()$ is an indicator function.
To incorporate the influence of the action $a_t$ on the future and account for the local greedy search, we use the discounted cumulative reward rather than the immediate reward to train the policy: 
\begin{equation}
\label{r}
R_{extr}(s_t,a_t) = \underbrace{r(s_{t},a_{t})}_{\text{immediate reward}} + \underbrace{\sum_{t'=t+1}^{T} \gamma^{t'-t}r(s_{t'},a_{t'})}_{\text{discounted future reward}}
\end{equation}
where $\gamma$ is the discounted factor (0.95 in our experiments). 

\paragraph{Intrinsic Reward} 
As discussed in Section~\ref{sec:critic}, we pre-train a matching critic to calculate the cycle-reconstruction intrinsic reward $R_{intr}$ (see Equation~\ref{eq:intrinsic}), promoting the alignment between the language instruction $\mathcal{X}$ and the trajectory $\tau$. It encourages the agent to respect the instruction and penalizes the paths that deviate from what the instruction indicates.

With both the extrinsic and intrinsic reward functions, the RL loss can be written as
\begin{equation}
    L_{rl} = - \mathbb{E}_{a_t \sim \pi_{\theta}}[A_t]
\end{equation}
where the advantage function $A_t = R_{extr} + \delta R_{intr}$. $\delta$ is a hyperparameter weighing the intrinsic reward. Based on REINFORCE algorithm~\cite{williams1992simple}, the gradient of non-differentiable, reward-based loss function can be derived as

\begin{equation}
    \nabla_{\theta} L_{rl} = - A_t \nabla_{\theta} \log \pi_{\theta}(a_t | s_t) 
\end{equation}


\begin{figure}
\begin{center}
\includegraphics[width=0.45\textwidth]{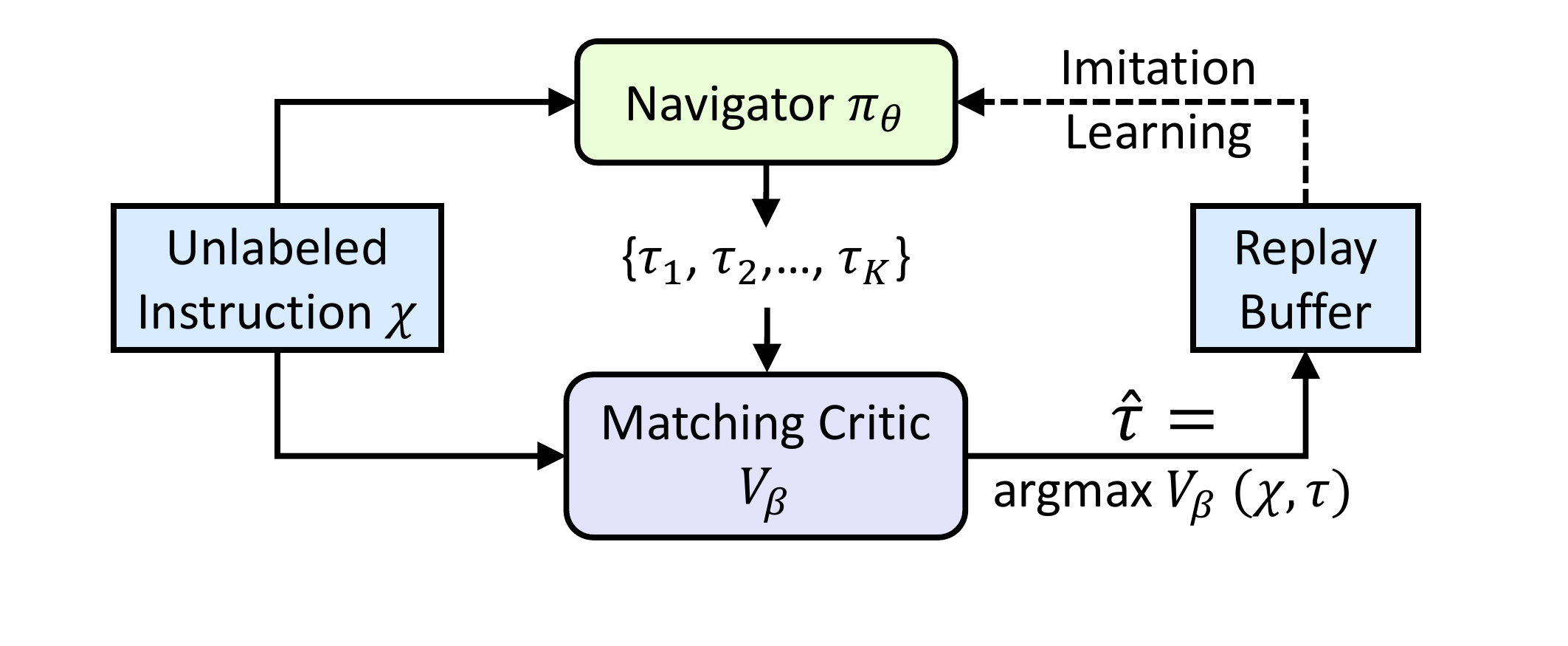}  
\end{center}
\caption{SIL for exploration on unlabeled data.}
\label{fig:sil}
\end{figure}

\section{Self-Supervised Imitation Learning}
The last section introduces the effective RCM method for generic vision-language navigation task, whose standard setting is to train the agent on seen environments and test it on unseen environments without exploration.
In this section we discuss a different setting where the agent is allowed to explore unseen environments without ground-truth demonstrations. This is of practical benefit because it facilitates lifelong learning and adaption to new environments. 

To this end, we propose a Self-Supervised Imitation Learning (SIL) method to imitate the agent's own past good decisions. As shown in Figure~\ref{fig:sil}, given a natural language instruction $\mathcal{X}$ without paired demonstrations and ground-truth target location, the navigator produces a set of possible trajectories and then stores the best trajectory $\hat{\tau}$ that is determined by matching critic $V_{\beta}$ into a replay buffer, in formula, 
\begin{equation}\label{eq:sil_argmax}
    \hat{\tau} = \arg \max_{\tau} V_{\beta}(\mathcal{X}, \tau)
\end{equation}
The matching critic evaluates the trajectories with the cycle-reconstruction reward as introduced in Section~\ref{sec:critic}.
Then by exploiting the good trajectories in the replay buffer, the agent is indeed optimizing the following objective with self-supervision. The target location is unknown and thus there is no supervision from the environment.
\begin{equation}
    L_{sil} = - R_{intr} \log \pi_{\theta}(a_t | s_t)
\end{equation}
Note that $L_{sil}$ can be viewed as the loss for policy gradient except that the off-policy Monte-Carlo return $R_{intr}$ is used instead of on-policy return. 
$L_{sil}$ can also be interpreted as the supervised learning loss with $\hat{\tau}$ as the ``ground truths'':
\begin{equation}
    L_{sil} = - \mathbb{E}[\log(\pi_{\theta}(\hat{a_t} | s_t))]
\end{equation}
where $\hat{a_t}$ is the action stored in the replay buffer using Equation~\ref{eq:sil_argmax}. Paired with a matching critic, the SIL method can be combined with various learning methods to approximate a better policy by imitating the previous best of itself. 


\section{Experiments and Analysis}

\subsection{Experimental Setup}

\paragraph{R2R Dataset}
We evaluate our approaches on the Room-to-Room (R2R) dataset~\cite{anderson2018vision} for vision-language navigation in real 3D environments, which is built
upon the Matterport3D dataset~\cite{chang2017matterport3d}. The R2R dataset has 7,189 paths that capture most of the visual diversity and 21,567 human-annotated instructions with an average length of 29 words. The R2R dataset is split into training, seen validation, unseen validation, and test sets. The seen validation set shares the same environments with the training set. While both the unseen validation and test sets contain distinct environments that do not appear in the other sets.

\paragraph{Testing Scenarios}
The standard testing scenario of the VLN task is to train the agent in seen environments and then test it in previously unseen environments in a zero-shot fashion. There is no prior exploration on the test set. This setting is preferred and able to clearly measure the generalizability of the navigation policy, so we evaluate our RCM approach under the standard testing scenario. 

Furthermore, exploration in unseen environments is certainly meaningful in practice, \eg, in-home robots are expected to explore and adapt to a new environment. So we introduce a lifelong learning scenario where the agent is encouraged to learn from trials and errors on the unseen environments. In this case, how to effectively explore the unseen validation or test set where there are no expert demonstrations becomes an important task to study. 
\paragraph{Evaluation Metrics}
We report five evaluation metrics as used by the VLN Challenge: Path Length (PL), Navigation Error (NE), Oracle Success Rate (OSR), Success Rate (SR), 
and Success rate weighted by inverse Path Length (SPL).\footnote{PL: the total length of the executed path. NE: the shortest-path distance between the agent's final position and the target. OSR: the success rate at the closest point to the goal that the agent has visited along the trajectory. SR: the percentage of predicted end-locations within 3m of the target locations. SPL: SPL trades-off Success Rate against Path Length, which is defined in \cite{anderson2018evaluation}.} 
Among those metrics, SPL is the recommended primary measure of navigation performance~\cite{anderson2018evaluation}, as it considers both effectiveness and efficiency. The other metrics are also reported as auxiliary measures.

\paragraph{Implementation Details}
Following prior work~\cite{anderson2018vision,wang2018look,fried2018speaker}, ResNet-152 CNN features~\cite{he2016deep} are extracted for all images without fine-tuning. The pretrained GloVe word embeddings~\cite{pennington2014glove} are used for initialization and then fine-tuned during training. We train the matching critic with human demonstrations and then fix it during policy learning. Then we warm start the policy via SL with a learning rate 1e-4, and then switch to RL training with a learning rate 1e-5 (same for SIL). Adam optimizer~\cite{kingma2014adam} is used to optimize all the parameters.
More details can be found in the appendix.

\begin{table}[t]
\small
\setlength{\tabcolsep}{4pt}
\begin{center}
  \begin{tabular}{ l c c c c c }
        \toprule
        \multicolumn{6}{c}{\textbf{Test Set (VLN Challenge Leaderboard)}} \\
        Model & PL $\downarrow$ & NE $\downarrow$ & OSR $\uparrow$ & SR $\uparrow$ & \textbf{SPL} $\uparrow$ \\
        \midrule
        Random 
        & 9.89 & 9.79 & 18.3 & 13.2 & 12 \\
        seq2seq~\cite{anderson2018vision}
        & \textbf{8.13} & 7.85 & 26.6 & 20.4 & 18 \\
        RPA~\cite{wang2018look}
        & 9.15 & 7.53 & 32.5 & 25.3 & 23 \\
        Speaker-Follower~\cite{fried2018speaker} 
        & 14.82 & 6.62 & 44.0 &    35.0 & 28 \\
        \quad $+$ beam search & \textit{\underline{1257.38}} & 4.87 & 96.0 & 53.5 & \textit{\underline{1}} \\
        \midrule
        \textbf{Ours} & \\
        RCM 
        & 15.22 & \textbf{6.01} & \textbf{50.8} & \textbf{43.1} & 35 \\ 
        RCM + SIL (train)
        & \textbf{11.97} & 6.12 & 49.5 & 43.0 & \textbf{38} \\ 
        \bottomrule
        RCM + SIL (unseen)\tablefootnote{The results of using SIL to explore unseen environments are only used to validate its effectiveness for lifelong learning, which is not directly comparable to other models due to different learning scenarios.}
        & 9.48 & 4.21 & 66.8 & 60.5 & 59 \\
        \bottomrule
    \end{tabular}
\end{center}
\caption{Comparison on the R2R test set~\cite{anderson2018vision}. 
Our RCM model significantly outperforms the SOTA methods, especially on SPL (the primary metric for navigation tasks~\cite{anderson2018evaluation}). Moreover, using SIL to imitate itself on the training set can further improve its efficiency: the path length is shortened by 3.25m. 
Note that with beam search, the agent executes $K$ trajectories at test time and chooses the most confident one as the ending point, which results in a super long path and is heavily penalized by SPL. }
\label{table:sota}
\end{table} 

\begin{table*}
\small
\begin{center}
  \begin{tabular}{ c l  c c c c  c c c c }
        \toprule
         & & \multicolumn{4}{c}{\textbf{Seen Validation}} & \multicolumn{4}{c}{\textbf{Unseen Validation}} \\
        \cmidrule(lr){3-6}\cmidrule(lr){7-10}
        \# & Model & \underline{PL} $\downarrow$ & NE $\downarrow$ & OSR $\uparrow$ & \underline{SR} $\uparrow$ & \underline{PL} $\downarrow$ & NE $\downarrow$ & OSR $\uparrow$ & \underline{SR} $\uparrow$ \\
        \midrule
        0 & Speaker-Follower (no beam search)~\cite{fried2018speaker}
        & - & 3.36 & 73.8 & 66.4 & - & 6.62 & 45.0 & 35.5 \\
        \midrule
        1 & RCM + SIL (train)
        & \underline{\textbf{10.65}} & 3.53 & 75.0 & 66.7 & \underline{\textbf{11.46}} & 6.09 & 50.1 &  \underline{\textbf{42.8}}
        \\
        2 & RCM
        & 11.92 & 3.37 & 76.6 & 67.4 & 14.84 & \textbf{5.88} & \textbf{51.9} & 42.5
        \\
        3 & \quad $-$ intrinsic reward 
        & 12.08 & 3.25 & \textbf{77.2} & \underline{\textbf{67.6}} & 15.00 & 6.02 & 50.5 & 40.6 
        \\ 
        4 & \quad $-$ extrinsic reward = pure SL
        & 11.99 & 3.22 & 76.7 & 66.9 & 14.83 & 6.29 & 46.5 & 37.7 
        \\
        5 & \quad $-$ cross-modal reasoning
        & 11.88 & \textbf{3.18} & 73.9 & 66.4 & 14.51 & 6.47 & 44.8 & 35.7
        \\
        \midrule
        6 & RCM $+$ SIL (unseen)
        & \textbf{10.13}    & \textbf{2.78} & \textbf{79.7} & \textbf{73.0} & \textbf{9.12} & \textbf{4.17} & \textbf{69.31} & \textbf{61.3} 
        \\
        \bottomrule
    \end{tabular}
\end{center}
\caption{Ablation study on seen and unseen validation sets. We report the performance of the speaker-follower model without beam search as the baseline. 
Row 1-5 shows the influence of each individual component by successively removing it from the final model. Row 6 illustrates the power of SIL on exploring unseen environments with self-supervision. Please see Section~\ref{sec:component} for more detailed analysis.
}
\label{table:ablation}
\end{table*} 

\subsection{Results on the Test Set}
\paragraph{Comparison with SOTA}
We compare the performance of RCM to the previous state-of-the-art (SOTA) methods on the test set of the R2R dataset, which is held out as the VLN Challenge. The results are shown in Table~\ref{table:sota}, where we compare RCM to a set of baselines: (1) \textit{Random}: randomly take a direction to move forward at each step until five steps.
(2) \textit{seq2seq}: the best-performing sequence-to-sequence model as reported in the original dataset paper~\cite{anderson2018vision}, which is trained with the student-forcing method.  
(3) \textit{RPA}: a reinforced planning-ahead model that combines model-free and model-based reinforcement learning for VLN~\cite{wang2018look}.
(4) \textit{Speaker-Follower}: a compositional Speaker-Follower method that combines data augmentation, panoramic action space, and beam search for VLN~\cite{fried2018speaker}. 

As can be seen in Table~\ref{table:sota}, RCM significantly outperforms the existing methods, improving the SPL score from 28\% to 35\%\footnote{Note that our RCM model also utilizes the panoramic action space and augmented data in \cite{fried2018speaker} for a fair comparison.}. The improvement is consistently observed on the other metrics, \eg, the success rate is increased by 8.1\%.
Moreover, using SIL to imitate the RCM agent's previous best behaviors on the training set can approximate a more efficient policy, whose average path length is reduced from 15.22m to 11.97m and which achieves the best result (38\%) on SPL. Therefore, we submit the results of \textit{RCM + SIL (train)} to the VLN Challenge, ranking first among prior work in terms of SPL.
It is worth noticing that beam search is not practical in reality, because it needs to execute a very long trajectory before making the decision, which is punished heavily by the primary metric SPL. So we are mainly comparing the results without beam search.

\paragraph{Self-Supervised Imitation Learning}
As mentioned above, for a standard VLN setting, we employ SIL on the training set to learn an efficient policy.  
For the lifelong learning scenario, we test the effectiveness of SIL on exploring unseen environments (the validation and test sets). 
It is noticeable in Table~\ref{table:sota} that SIL indeed leads to a better policy even without knowing the target locations. SIL improves RCM by 17.5\% on SR and 21\% on SPL. 
Similarly, the agent also learns a more efficient policy that takes less number of steps (the average path length is reduced from 15.22m to 9.48m) but obtains a higher success rate.
The key difference between SIL and beam search is that SIL optimizes the policy itself by play-and-imitate while beam search only makes a greedy selection of the rollouts of the existing policy.  
But we would like to point out that due to different learning scenarios, the results of \textit{RCM + SIL (unseen)} cannot be directly compared with other methods following the standard settings of the VLN challenge. 

\subsection{Ablation Study}
\paragraph{Effect of Individual Components}\label{sec:component}
We conduct an ablation study to illustrate the effect of each component on both seen and unseen validation sets in Table~\ref{table:ablation}. Comparing Row 1 and Row 2, we observe the efficiency of the learned policy by imitating the best of itself on the training set. 
Then we start with the RCM model in Row 2, and successively remove the \textit{intrinsic reward}, \textit{extrinsic reward}, and \textit{cross-modal reasoning} to demonstrate their importance. 

Removing the intrinsic reward (Row 3), we notice that the success rate (SR) on unseen environments drops 1.9 points while it is almost fixed on seen environments (0.2$\uparrow$). It evaluates the alignment between instructions and trajectories, serving as a complementary supervision besides of the feedback from the environment, therefore it works better for the unseen environments that require more supervision due to lack of exploration. 
This also indirectly validates the importance of exploration on unseen environments.

Furthermore, the results of Row 4 (the RCM model with only supervised learning) validate the superiority of reinforcement learning compared to purely supervised learning on the VLN task. Meanwhile, since eventually the results are evaluated based on the success rate (SR) and path length (PL), directly optimizing the extrinsic reward signals can guarantee the stability of the reinforcement learning and bring a big performance gain. 

\begin{figure*}
\centering
\includegraphics[width=0.9\textwidth]{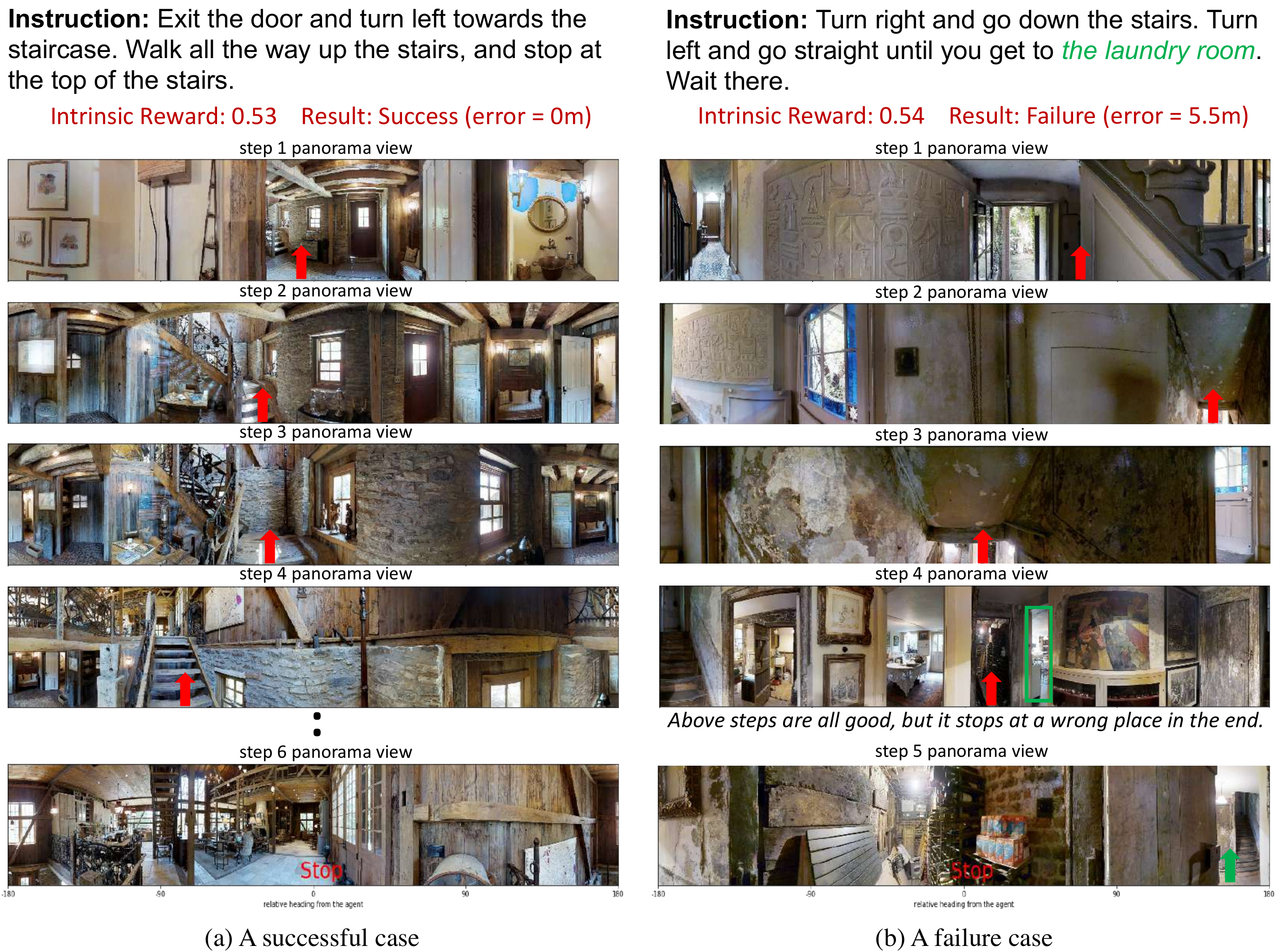} 
\caption{Qualitative examples from the unseen validation set. }
\label{fig:demo}
\end{figure*}

We then verify the strength of our cross-modal reasoning navigator by comparing it (Row 4) with an attention-based sequence-to-sequence model (Row 5) that utilizes the previous hidden state $h_{t-1}$ to attend to both the visual and textual features at decoding time. Everything else is exactly the same except the cross-modal attention design. Evidently, our navigator improves upon the baseline by considering history context, visually-conditioned textual context, and textually-conditioned visual context for decision making. 

In the end, we demonstrate the effectiveness of the proposed SIL method for exploration in Row 6. Considerable performance boosts have been obtained on both seen and unseen environments, as the agent learns how to better execute the instructions from its own previous experience. 

\paragraph{Generalizability}\label{sec:generalize} 
Another observation from the experiments (\eg, see Table~\ref{table:ablation}) is that our RCM approach is much more generalizable to unseen environments compared with the baseline. The improvements on the seen and unseen validation sets are 0.3 and 7.1 points, respectively. So is the SIL method, which explicitly explores the unseen environments and tremendously reduces the success rate performance gap between seen and unseen environments from 30.7\% (Row 5) to 11.7\% (Row 6).  

\paragraph{Qualitative Analysis}
For a more intuitive view of how our model works for the VLN task, we visualize two qualitative examples in Figure~\ref{fig:demo}. Particularly, we choose two examples, both with high intrinsic rewards. In (a), the agent successfully reaches the target destination, with a comprehensive understanding of the natural language instruction. While in (b), the intrinsic reward is also high, which indicates most of the agent's actions are good, but it is also noticeable that the agent fails to recognize \textit{the laundry room} at the end of the trajectory, which shows the importance of more precise visual grounding in the navigation task. 

\section{Conclusion}
In this paper we present two novel approaches, RCM and SIL, which combine the strength of reinforcement learning and self-supervised imitation learning for the vision-language navigation task. Experiments illustrate the effectiveness and efficiency of our methods under both the standard testing scenario and the lifelong learning scenario. 
Moreover, our methods show strong generalizability in unseen environments. 
The proposed learning frameworks are modular and model-agnostic, which allow the components to be improved separately. 
We also believe that the idea of learning more fine-grained intrinsic rewards, in addition to the coarse external signals, is commonly applicable to various embodied agent tasks, and the idea SIL can be generally adopted to explore other unseen environments.

\section*{Acknowledgment}
This work was partly performed when the first author was interning at Microsoft Research. The authors thank Peter Anderson and Pengchuan Zhang for their helpful discussions, and Ronghang Hu for his visualization code. 

{\small
\bibliographystyle{ieee}
\bibliography{egbib}

\begin{thebibliography}{10}\itemsep=-1pt

\bibitem{anderson2018evaluation}
P.~Anderson, A.~Chang, D.~S. Chaplot, A.~Dosovitskiy, S.~Gupta, V.~Koltun,
  J.~Kosecka, J.~Malik, R.~Mottaghi, M.~Savva, et~al.
\newblock On evaluation of embodied navigation agents.
\newblock {\em arXiv preprint arXiv:1807.06757}, 2018.

\bibitem{anderson2018bottom}
P.~Anderson, X.~He, C.~Buehler, D.~Teney, M.~Johnson, S.~Gould, and L.~Zhang.
\newblock Bottom-up and top-down attention for image captioning and visual
  question answering.
\newblock In {\em CVPR}, volume~3, page~6, 2018.

\bibitem{anderson2018vision}
P.~Anderson, Q.~Wu, D.~Teney, J.~Bruce, M.~Johnson, N.~S{\"u}nderhauf, I.~Reid,
  S.~Gould, and A.~van~den Hengel.
\newblock Vision-and-language navigation: Interpreting visually-grounded
  navigation instructions in real environments.
\newblock In {\em Proceedings of the IEEE Conference on Computer Vision and
  Pattern Recognition (CVPR)}, volume~2, 2018.

\bibitem{VQA}
S.~Antol, A.~Agrawal, J.~Lu, M.~Mitchell, D.~Batra, C.~L. Zitnick, and
  D.~Parikh.
\newblock {VQA}: {V}isual {Q}uestion {A}nswering.
\newblock In {\em International Conference on Computer Vision (ICCV)}, 2015.

\bibitem{bellemare2016unifying}
M.~Bellemare, S.~Srinivasan, G.~Ostrovski, T.~Schaul, D.~Saxton, and R.~Munos.
\newblock Unifying count-based exploration and intrinsic motivation.
\newblock In {\em Advances in Neural Information Processing Systems}, pages
  1471--1479, 2016.

\bibitem{chang2017matterport3d}
A.~Chang, A.~Dai, T.~Funkhouser, M.~Halber, M.~Nie{\ss}ner, M.~Savva, S.~Song,
  A.~Zeng, and Y.~Zhang.
\newblock Matterport3d: Learning from rgb-d data in indoor environments.
\newblock {\em arXiv preprint arXiv:1709.06158}, 2017.

\bibitem{chen2015mind}
X.~Chen and C.~Lawrence~Zitnick.
\newblock Mind's eye: A recurrent visual representation for image caption
  generation.
\newblock In {\em Proceedings of the IEEE conference on computer vision and
  pattern recognition}, pages 2422--2431, 2015.

\bibitem{embodiedqa}
A.~Das, S.~Datta, G.~Gkioxari, S.~Lee, D.~Parikh, and D.~Batra.
\newblock {E}mbodied {Q}uestion {A}nswering.
\newblock In {\em Proceedings of the IEEE Conference on Computer Vision and
  Pattern Recognition (CVPR)}, 2018.

\bibitem{visdial}
A.~Das, S.~Kottur, K.~Gupta, A.~Singh, D.~Yadav, J.~M. Moura, D.~Parikh, and
  D.~Batra.
\newblock {V}isual {D}ialog.
\newblock In {\em Proceedings of the IEEE Conference on Computer Vision and
  Pattern Recognition (CVPR)}, 2017.

\bibitem{imagenet_cvpr09}
J.~Deng, W.~Dong, R.~Socher, L.-J. Li, K.~Li, and L.~Fei-Fei.
\newblock {ImageNet: A Large-Scale Hierarchical Image Database}.
\newblock In {\em CVPR09}, 2009.

\bibitem{lrcn2014}
J.~Donahue, L.~A. Hendricks, S.~Guadarrama, M.~Rohrbach, S.~Venugopalan,
  K.~Saenko, and T.~Darrell.
\newblock Long-term recurrent convolutional networks for visual recognition and
  description.
\newblock In {\em CVPR}, 2015.

\bibitem{fang2015captions}
H.~Fang, S.~Gupta, F.~Iandola, R.~K. Srivastava, L.~Deng, P.~Doll{\'a}r,
  J.~Gao, X.~He, M.~Mitchell, J.~C. Platt, et~al.
\newblock From captions to visual concepts and back.
\newblock In {\em Proceedings of the IEEE conference on computer vision and
  pattern recognition}, pages 1473--1482, 2015.

\bibitem{fried2018speaker}
D.~Fried, R.~Hu, V.~Cirik, A.~Rohrbach, J.~Andreas, L.-P. Morency,
  T.~Berg-Kirkpatrick, K.~Saenko, D.~Klein, and T.~Darrell.
\newblock Speaker-follower models for vision-and-language navigation.
\newblock In {\em Advances in Neural Information Processing Systems (NIPS)},
  2018.

\bibitem{gaosurvey}
J.~Gao, M.~Galley, and L.~Li.
\newblock Neural approaches to conversational ai.
\newblock {\em arXiv preprint arXiv:1809.08267}, 2018.

\bibitem{he2016deep}
K.~He, X.~Zhang, S.~Ren, and J.~Sun.
\newblock Deep residual learning for image recognition.
\newblock In {\em Proceedings of the IEEE conference on computer vision and
  pattern recognition}, pages 770--778, 2016.

\bibitem{hemachandra2015learning}
S.~Hemachandra, F.~Duvallet, T.~M. Howard, N.~Roy, A.~Stentz, and M.~R. Walter.
\newblock Learning models for following natural language directions in unknown
  environments.
\newblock {\em arXiv preprint arXiv:1503.05079}, 2015.

\bibitem{hochreiter1997long}
S.~Hochreiter and J.~Schmidhuber.
\newblock Long short-term memory.
\newblock {\em Neural computation}, 9(8):1735--1780, 1997.

\bibitem{houthooft2016vime}
R.~Houthooft, X.~Chen, Y.~Duan, J.~Schulman, F.~De~Turck, and P.~Abbeel.
\newblock Vime: Variational information maximizing exploration.
\newblock In {\em Advances in Neural Information Processing Systems}, pages
  1109--1117, 2016.

\bibitem{hu2016segmentation}
R.~Hu, M.~Rohrbach, and T.~Darrell.
\newblock Segmentation from natural language expressions.
\newblock In {\em European Conference on Computer Vision}, pages 108--124.
  Springer, 2016.

\bibitem{hu2016natural}
R.~Hu, H.~Xu, M.~Rohrbach, J.~Feng, K.~Saenko, and T.~Darrell.
\newblock Natural language object retrieval.
\newblock In {\em Proceedings of the IEEE Conference on Computer Vision and
  Pattern Recognition}, pages 4555--4564, 2016.

\bibitem{huang2018turbo}
Q.~Huang, P.~Zhang, D.~Wu, and L.~Zhang.
\newblock Turbo learning for captionbot and drawingbot.
\newblock In {\em Advances in Neural Information Processing Systems (NIPS)},
  2018.

\bibitem{karpathy2015deep}
A.~Karpathy and L.~Fei-Fei.
\newblock Deep visual-semantic alignments for generating image descriptions.
\newblock In {\em Proceedings of the IEEE conference on computer vision and
  pattern recognition}, pages 3128--3137, 2015.

\bibitem{ke2017tactile}
L.~Ke, X.~Li, Y.~Bisk, A.~Holtzman, Z.~Gan, J.~Liu, J.~Gao, Y.~Choi, and
  S.~Srinivasa.
\newblock Tactical rewind: Self-correction via backtracking in
  vision-and-language navigation.
\newblock {\em arXiv preprint arXiv:1903.02547}, 2019.

\bibitem{kingma2014adam}
D.~P. Kingma and J.~Ba.
\newblock Adam: A method for stochastic optimization.
\newblock {\em arXiv preprint arXiv:1412.6980}, 2014.

\bibitem{kolve2017ai2}
E.~Kolve, R.~Mottaghi, D.~Gordon, Y.~Zhu, A.~Gupta, and A.~Farhadi.
\newblock Ai2-thor: An interactive 3d environment for visual ai.
\newblock {\em arXiv preprint arXiv:1712.05474}, 2017.

\bibitem{lipton2016efficient}
Z.~C. Lipton, J.~Gao, L.~Li, X.~Li, F.~Ahmed, and L.~Deng.
\newblock Efficient exploration for dialogue policy learning with bbq networks
  \& replay buffer spiking.
\newblock {\em arXiv preprint arXiv:1608.05081}, 2016.

\bibitem{ma2019self}
C.-Y. Ma, J.~Lu, Z.~Wu, G.~AlRegib, Z.~Kira, R.~Socher, and C.~Xiong.
\newblock Self-monitoring navigation agent via auxiliary progress estimation.
\newblock {\em arXiv preprint arXiv:1901.03035}, 2019.

\bibitem{ma2019regretful}
C.-Y. Ma, Z.~Wu, G.~AlRegib, C.~Xiong, and Z.~Kira.
\newblock The regretful agent: Heuristic-aided navigation through progress
  estimation.
\newblock {\em arXiv preprint arXiv:1903.01602}, 2019.

\bibitem{mirowski2016learning}
P.~Mirowski, R.~Pascanu, F.~Viola, H.~Soyer, A.~J. Ballard, A.~Banino,
  M.~Denil, R.~Goroshin, L.~Sifre, K.~Kavukcuoglu, et~al.
\newblock Learning to navigate in complex environments.
\newblock {\em arXiv preprint arXiv:1611.03673}, 2016.

\bibitem{mousavian2018visual}
A.~Mousavian, A.~Toshev, M.~Fiser, J.~Kosecka, and J.~Davidson.
\newblock Visual representations for semantic target driven navigation.
\newblock {\em arXiv preprint arXiv:1805.06066}, 2018.

\bibitem{nguyen2018vision}
K.~Nguyen, D.~Dey, C.~Brockett, and B.~Dolan.
\newblock Vision-based navigation with language-based assistance via imitation
  learning with indirect intervention.
\newblock {\em arXiv preprint arXiv:1812.04155}, 2018.

\bibitem{oh2018self}
J.~Oh, Y.~Guo, S.~Singh, and H.~Lee.
\newblock Self-imitation learning.
\newblock {\em arXiv preprint arXiv:1806.05635}, 2018.

\bibitem{ostrovski2017count}
G.~Ostrovski, M.~G. Bellemare, A.~v.~d. Oord, and R.~Munos.
\newblock Count-based exploration with neural density models.
\newblock {\em arXiv preprint arXiv:1703.01310}, 2017.

\bibitem{pathak2017curiosity}
D.~Pathak, P.~Agrawal, A.~A. Efros, and T.~Darrell.
\newblock Curiosity-driven exploration by self-supervised prediction.
\newblock In {\em International Conference on Machine Learning (ICML)}, volume
  2017, 2017.

\bibitem{pennington2014glove}
J.~Pennington, R.~Socher, and C.~Manning.
\newblock Glove: Global vectors for word representation.
\newblock In {\em Proceedings of the 2014 conference on empirical methods in
  natural language processing (EMNLP)}, pages 1532--1543, 2014.

\bibitem{elmo}
M.~E. Peters, M.~Neumann, M.~Iyyer, M.~Gardner, C.~Clark, K.~Lee, and
  L.~Zettlemoyer.
\newblock Deep contextualized word representations.
\newblock In {\em Proc. of NAACL}, 2018.

\bibitem{plummer2015flickr30k}
B.~A. Plummer, L.~Wang, C.~M. Cervantes, J.~C. Caicedo, J.~Hockenmaier, and
  S.~Lazebnik.
\newblock Flickr30k entities: Collecting region-to-phrase correspondences for
  richer image-to-sentence models.
\newblock In {\em Proceedings of the IEEE international conference on computer
  vision}, pages 2641--2649, 2015.

\bibitem{minos}
M.~Savva, A.~X. Chang, A.~Dosovitskiy, T.~Funkhouser, and V.~Koltun.
\newblock Minos: Multimodal indoor simulator for navigation in complex
  environments.
\newblock {\em arXiv preprint arXiv:1712.03931}, 2017.

\bibitem{schmidhuber1991adaptive}
J.~Schmidhuber.
\newblock Adaptive confidence and adaptive curiosity.
\newblock In {\em Institut fur Informatik, Technische Universitat Munchen,
  Arcisstr. 21, 800 Munchen 2}. Citeseer, 1991.

\bibitem{song2016ssc}
S.~Song, F.~Yu, A.~Zeng, A.~X. Chang, M.~Savva, and T.~Funkhouser.
\newblock Semantic scene completion from a single depth image.
\newblock {\em IEEE Conference on Computer Vision and Pattern Recognition},
  2017.

\bibitem{strehl2008analysis}
A.~L. Strehl and M.~L. Littman.
\newblock An analysis of model-based interval estimation for markov decision
  processes.
\newblock {\em Journal of Computer and System Sciences}, 74(8):1309--1331,
  2008.

\bibitem{tang2017exploration}
H.~Tang, R.~Houthooft, D.~Foote, A.~Stooke, O.~X. Chen, Y.~Duan, J.~Schulman,
  F.~DeTurck, and P.~Abbeel.
\newblock \# exploration: A study of count-based exploration for deep
  reinforcement learning.
\newblock In {\em Advances in Neural Information Processing Systems}, pages
  2753--2762, 2017.

\bibitem{tapaswi2016movieqa}
M.~Tapaswi, Y.~Zhu, R.~Stiefelhagen, A.~Torralba, R.~Urtasun, and S.~Fidler.
\newblock Movieqa: Understanding stories in movies through question-answering.
\newblock In {\em Proceedings of the IEEE conference on computer vision and
  pattern recognition}, pages 4631--4640, 2016.

\bibitem{thomason2018shifting}
J.~Thomason, D.~Gordan, and Y.~Bisk.
\newblock Shifting the baseline: Single modality performance on visual
  navigation \& qa.
\newblock {\em arXiv preprint arXiv:1811.00613}, 2018.

\bibitem{vaswani2017attention}
A.~Vaswani, N.~Shazeer, N.~Parmar, J.~Uszkoreit, L.~Jones, A.~N. Gomez,
  {\L}.~Kaiser, and I.~Polosukhin.
\newblock Attention is all you need.
\newblock In {\em Advances in Neural Information Processing Systems}, pages
  5998--6008, 2017.

\bibitem{vinyals2015show}
O.~Vinyals, A.~Toshev, S.~Bengio, and D.~Erhan.
\newblock Show and tell: A neural image caption generator.
\newblock In {\em Computer Vision and Pattern Recognition (CVPR), 2015 IEEE
  Conference on}, pages 3156--3164. IEEE, 2015.

\bibitem{wang2018AREL}
X.~Wang, W.~Chen, Y.-F. Wang, and W.~Y. Wang.
\newblock No metrics are perfect: Adversarial reward learning for visual
  storytelling.
\newblock In {\em Proceedings of the 56th Annual Meeting of the Association for
  Computational Linguistics (Volume 1: Long Papers)}, 2018.

\bibitem{wang2018video}
X.~Wang, W.~Chen, J.~Wu, Y.-F. Wang, and W.~Y. Wang.
\newblock Video captioning via hierarchical reinforcement learning.
\newblock In {\em The IEEE Conference on Computer Vision and Pattern
  Recognition (CVPR)}, 2018.

\bibitem{wang2018watch}
X.~Wang, Y.-F. Wang, and W.~Y. Wang.
\newblock Watch, listen, and describe: Globally and locally aligned cross-modal
  attentions for video captioning.
\newblock In {\em Proceedings of the 2018 Conference of the North American
  Chapter of the Association for Computational Linguistics: Human Language
  Technologies, Volume 2 (Short Papers)}, 2018.

\bibitem{wang2018look}
X.~Wang, W.~Xiong, H.~Wang, and W.~Y. Wang.
\newblock Look before you leap: Bridging model-free and model-based
  reinforcement learning for planned-ahead vision-and-language navigation.
\newblock In {\em The European Conference on Computer Vision (ECCV)}, September
  2018.

\bibitem{williams1992simple}
R.~J. Williams.
\newblock Simple statistical gradient-following algorithms for connectionist
  reinforcement learning.
\newblock {\em Machine learning}, 8(3-4):229--256, 1992.

\bibitem{xiazamirhe2018gibsonenv}
F.~Xia, A.~R.~Zamir, Z.-Y. He, A.~Sax, J.~Malik, and S.~Savarese.
\newblock Gibson {Env}: real-world perception for embodied agents.
\newblock In {\em Computer Vision and Pattern Recognition (CVPR), 2018 IEEE
  Conference on}. IEEE, 2018.

\bibitem{xu2015show}
K.~Xu, J.~Ba, R.~Kiros, K.~Cho, A.~Courville, R.~Salakhudinov, R.~Zemel, and
  Y.~Bengio.
\newblock Show, attend and tell: Neural image caption generation with visual
  attention.
\newblock In {\em International Conference on Machine Learning}, pages
  2048--2057, 2015.

\bibitem{yang2016stacked}
Z.~Yang, X.~He, J.~Gao, L.~Deng, and A.~Smola.
\newblock Stacked attention networks for image question answering.
\newblock In {\em Proceedings of the IEEE Conference on Computer Vision and
  Pattern Recognition}, pages 21--29, 2016.

\bibitem{yu2016video}
H.~Yu, J.~Wang, Z.~Huang, Y.~Yang, and W.~Xu.
\newblock Video paragraph captioning using hierarchical recurrent neural
  networks.
\newblock In {\em Proceedings of the IEEE conference on computer vision and
  pattern recognition}, pages 4584--4593, 2016.

\bibitem{yu2016modeling}
L.~Yu, P.~Poirson, S.~Yang, A.~C. Berg, and T.~L. Berg.
\newblock Modeling context in referring expressions.
\newblock In {\em European Conference on Computer Vision}, pages 69--85.
  Springer, 2016.

\bibitem{zhang2018man}
D.~Zhang, X.~Dai, X.~Wang, Y.-F. Wang, and L.~S. Davis.
\newblock Man: Moment alignment network for natural language moment retrieval
  via iterative graph adjustment.
\newblock {\em arXiv preprint arXiv:1812.00087}, 2018.

\bibitem{zhu2017target}
Y.~Zhu, R.~Mottaghi, E.~Kolve, J.~J. Lim, A.~Gupta, L.~Fei-Fei, and A.~Farhadi.
\newblock Target-driven visual navigation in indoor scenes using deep
  reinforcement learning.
\newblock In {\em Robotics and Automation (ICRA), 2017 IEEE International
  Conference on}, pages 3357--3364. IEEE, 2017.

\end{thebibliography}
}

\clearpage

\appendix
\section*{Supplementary Material}
\section{Training Details}
Following prior work~\cite{anderson2018vision,wang2018look,fried2018speaker}, ResNet-152 CNN features~\cite{he2016deep} are extracted for all images without fine-tuning. The pretrained GloVe word embeddings~\cite{pennington2014glove} are used for initialization and then fine-tuned during training. All the hyper-parameters are tuned on the validation sets.
We adopt the panoramic action space~\cite{fried2018speaker} where the action is to choose a navigable direction from the possible candidates. We set the maximal length of the action path as 10. The maximum length of the instruction is set as 80 and longer instructions are truncated.  
We train the matching critic with a learning rate 1e-4 and then fix it during policy learning. Then we warm start the policy via supervised learning loss with a learning rate 1e-4, and then switch to RL training with a learning rate 1e-5. 
Self-supervised imitation learning can be performed to further improve the policy: during the first epoch of SIL, the loaded policy produces 10 trajectories, of which the one with the highest intrinsic reward is stored in the replay buffer; those saved trajectories are then utilized to fine-tune the policy for a fixed number of iterations (the learning rate is 1e-5). 
Early stopping is used for all the training and Adam optimizer~\cite{kingma2014adam} is used to optimize all the parameters.
To avoid overfitting, we use an L2 weight decay of 0.0005 and a dropout ratio
of 0.5. The discounted factor $\gamma$ of our cumulative reward is 0.95. The weight $\sigma$ of the intrinsic reward is set as 2.

\section{Network Architecture}
\paragraph{Reasoning Navigator}
The language encoder consists of an LSTM with hidden size 512 and a word embedding layer of size 300. 
The inner dimensions of the three attention modules used to compute the history context, the textual context, and the visual context are 256, 512, and 256 respectively. The trajectory encoder is an LSTM with hidden size 512. The action embedding is a concatenation of the visual appearance feature vector of size 2048 and the orientation feature vector of size 128 (the 4-dimensional orientation feature $[sin \psi; cos \psi; sin \omega; cos \omega]$ are tiled 32 times as used in \cite{fried2018speaker}). The action predictor is composed of three weight matrices: the projection dimensions of $W_c$ and $W_u$ are both 256, and then an output layer $W_o$ together with a softmax layer are followed to obtain the probabilities over the possible navigable directions. 

\begin{figure}
\centering
\subfloat[Seen Validation]{{\includegraphics[width=0.24\textwidth]{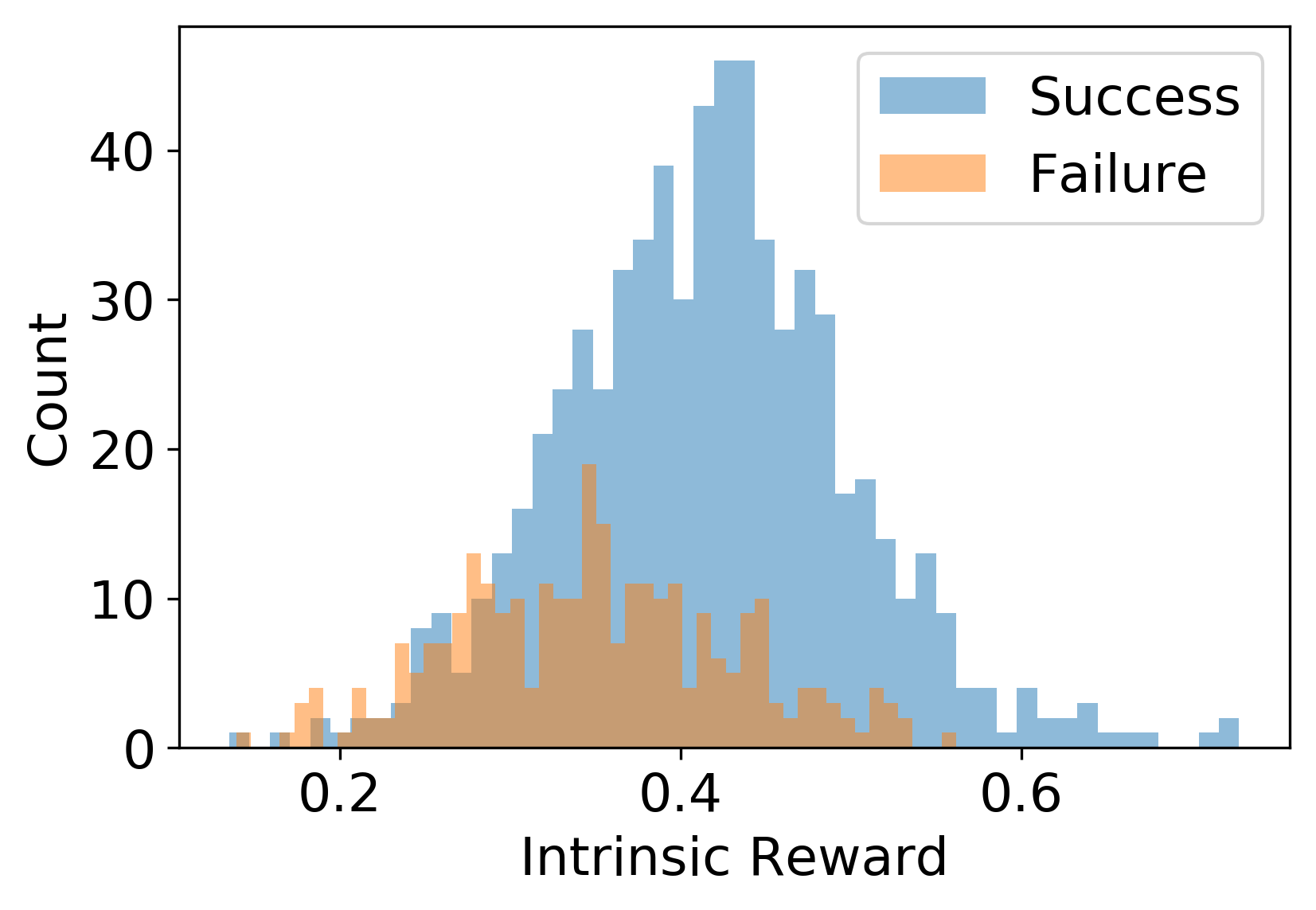} }}%
\subfloat[Unseen Validation]{{\includegraphics[width=0.24\textwidth]{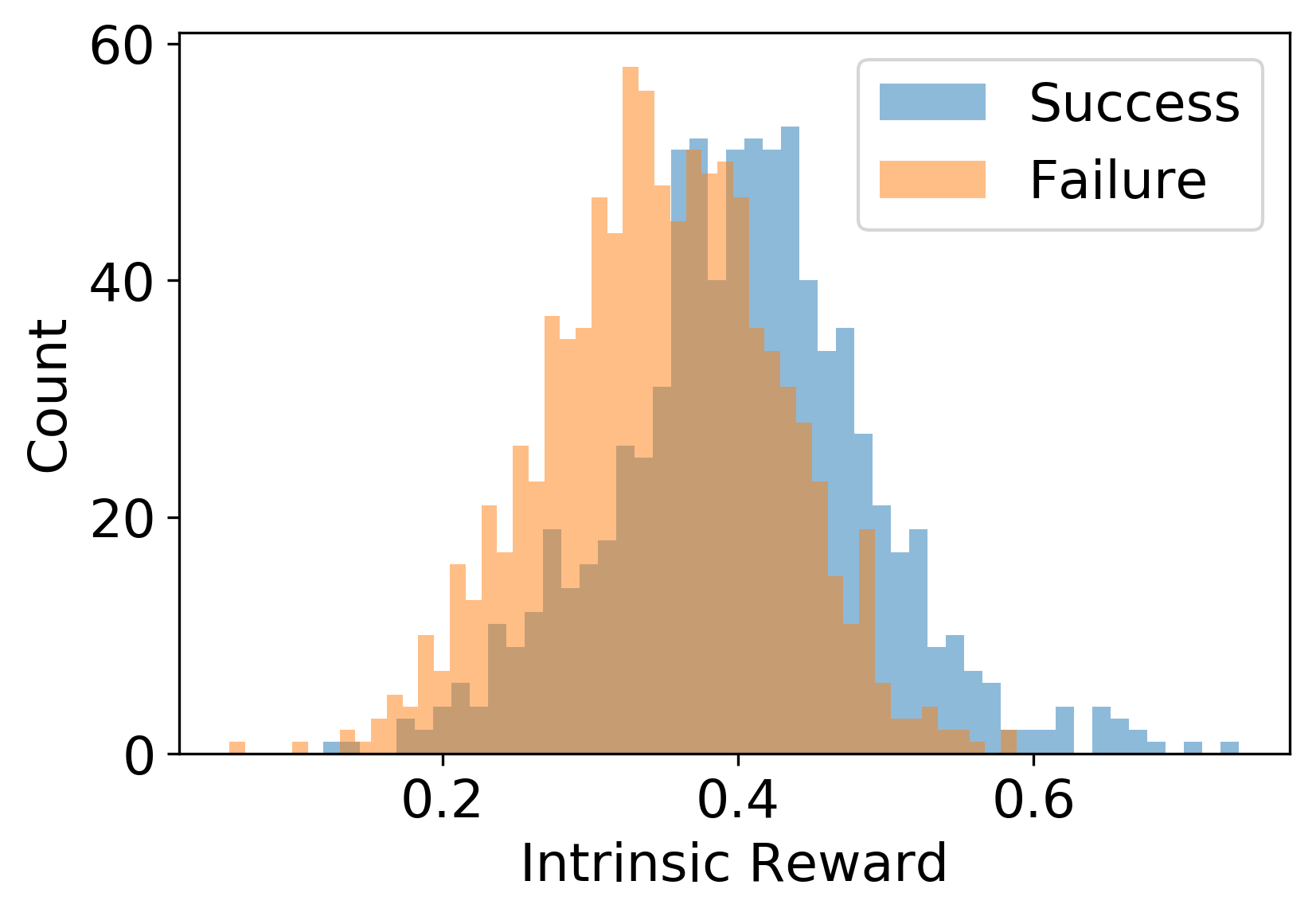} }}%
\caption{Visualization of the intrinsic reward on seen and unseen validation sets.}
\label{fig:vis}
\end{figure}

\begin{figure}
\centering
\includegraphics[width=0.5\textwidth]{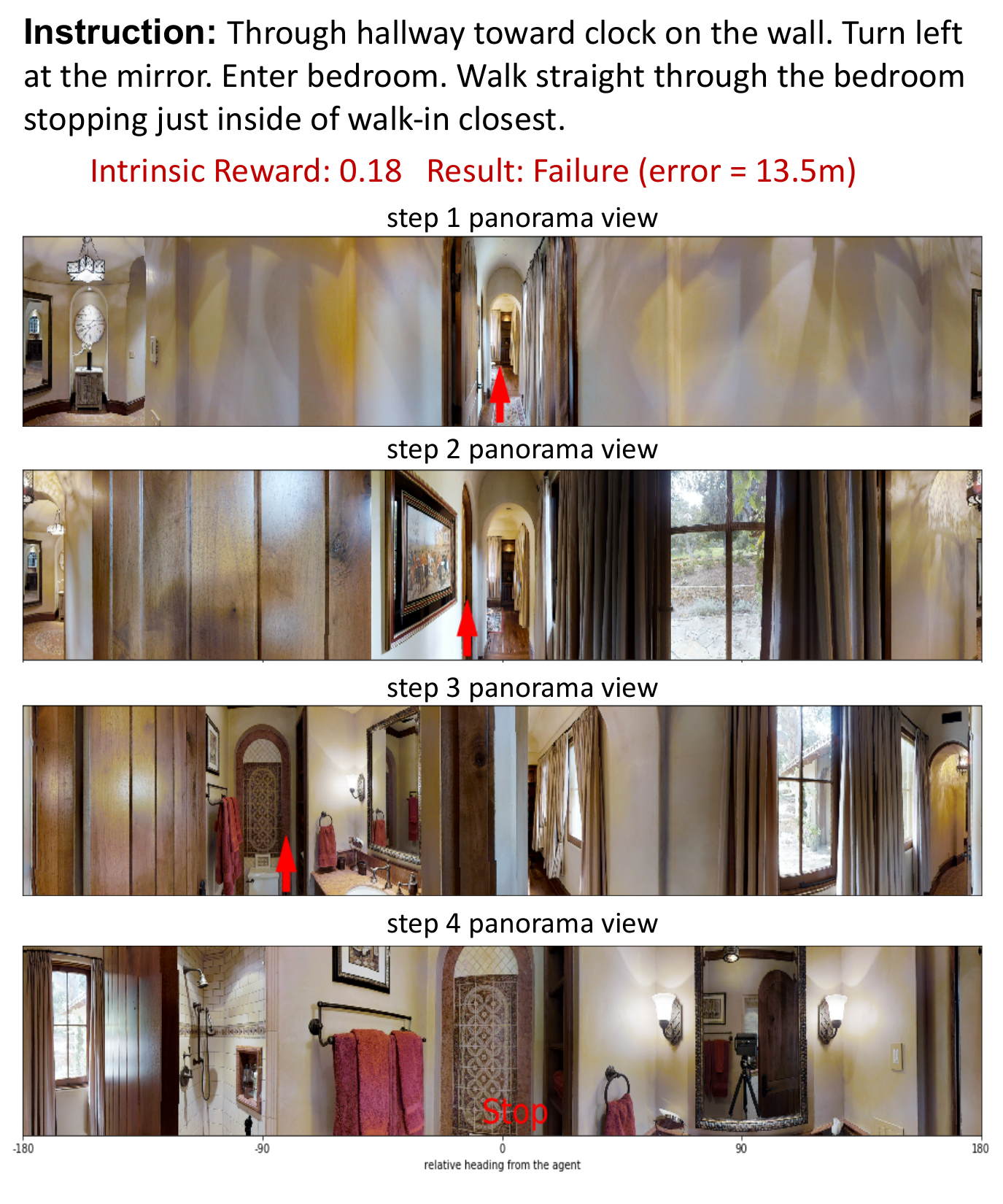} 
\caption{Misunderstanding of the instruction.}
\label{fig:neg_short}
\end{figure}

\begin{figure*}
\centering
\includegraphics[width=1\textwidth]{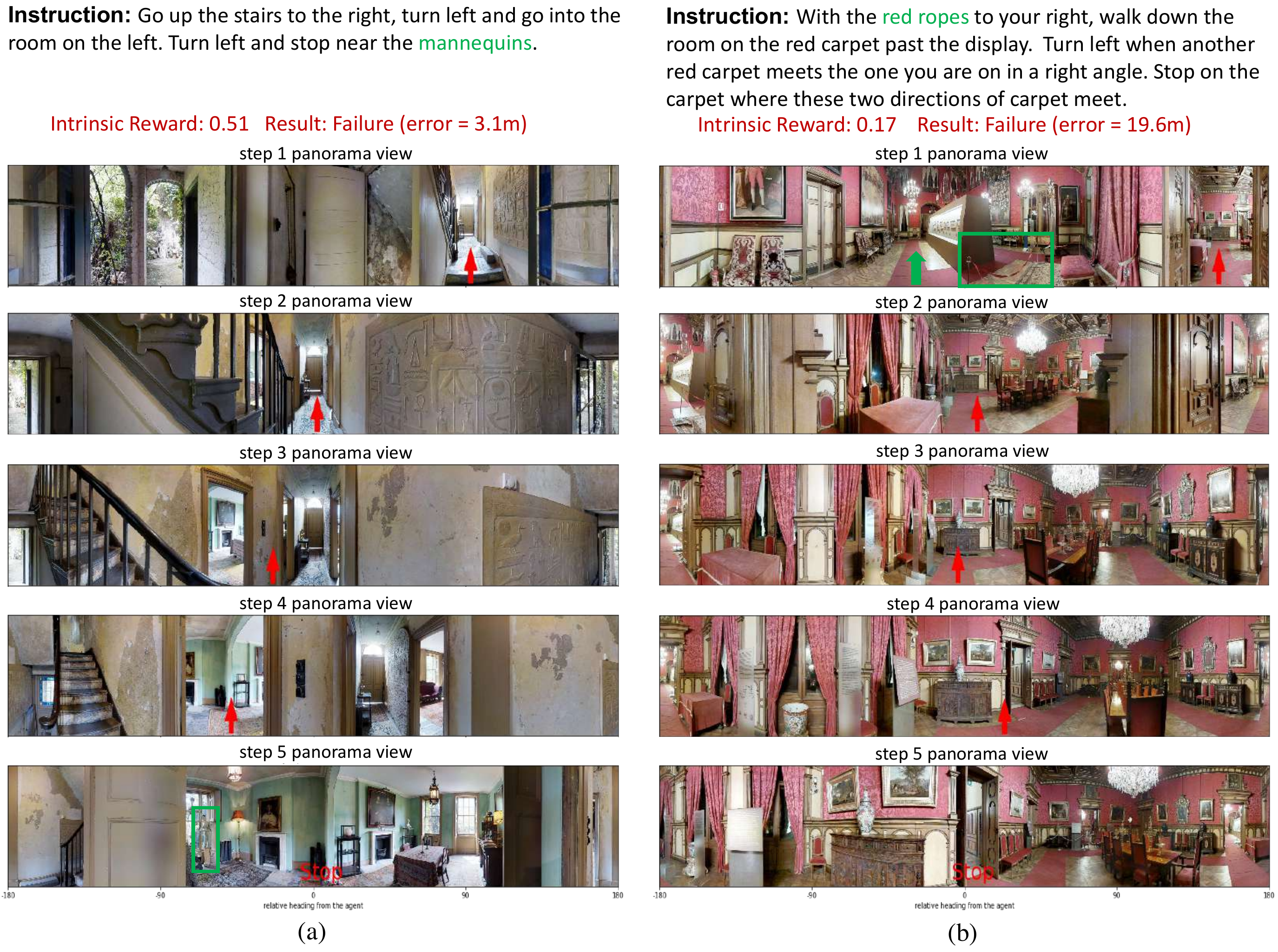} 
\caption{Ground errors where objects were not recognized from the visual scene.}
\label{fig:neg_v}
\end{figure*}

\begin{figure}
\centering
\includegraphics[width=0.5\textwidth]{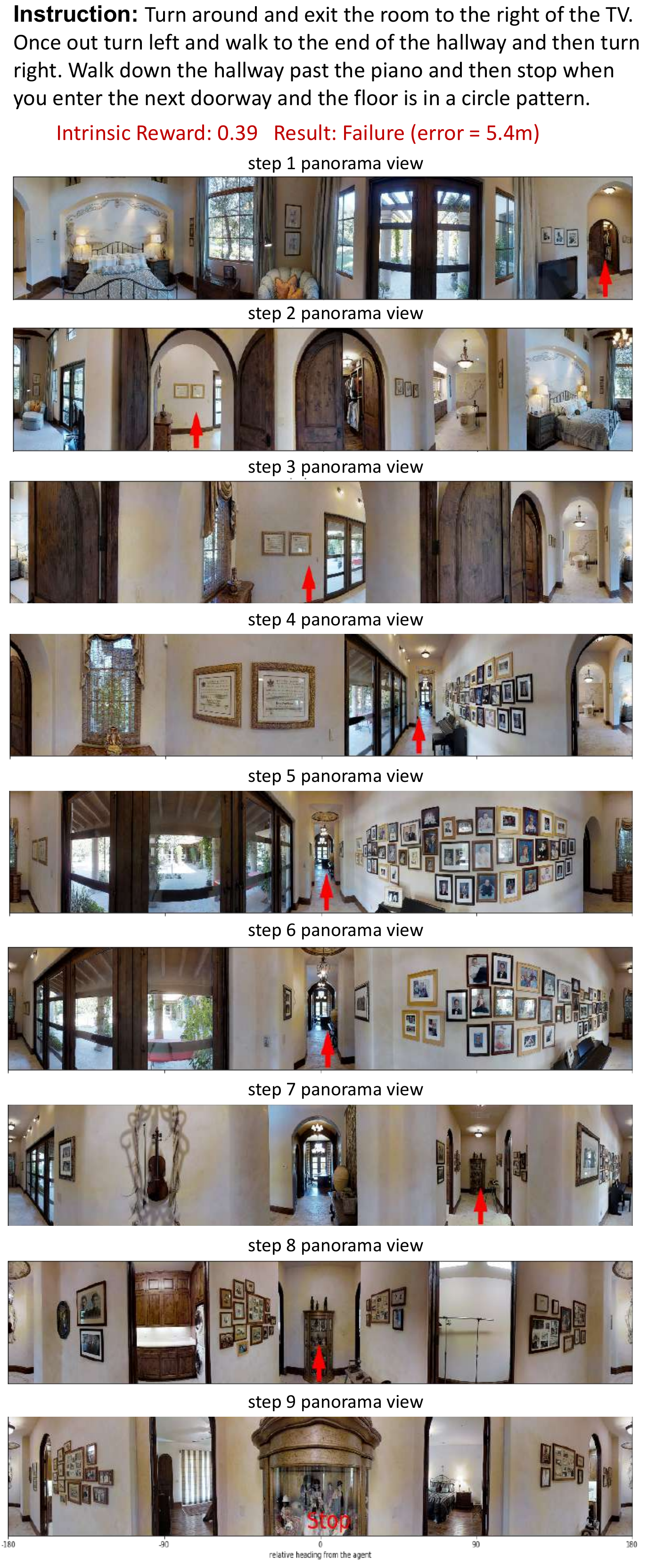} 
\caption{Failure of executing a relatively complicated instruction. }
\label{fig:neg_long}
\end{figure}

\paragraph{Matching Critic}
The matching critic consists of an attention-based trajectory encoder with the same architecture as the one in the navigator, its own word embedding layer of size 300, and an attention-based language decoder. The language decoder is composed of an attention module (whose projection dimension is 512) over the encoded features, an LSTM of hidden size 512, and a multi-layer perceptron (Linear $\rightarrow$ Tanh $\rightarrow$ Linear $\rightarrow$ SoftMax) that converts the hidden state into probabilities of all the words in the vocabulary. 

\section{Visualizing Intrinsic Reward}
In Figure~\ref{fig:vis}, we plot the histogram distributions of the intrinsic rewards (produced by our submitted model) on both seen and unseen validation sets. On the one hand, the intrinsic reward is aligned with the success rate to some extent, because the successful examples are receiving higher averaged intrinsic rewards than the failed ones. On the other hand, the complementary intrinsic reward provides more fine-grained reward signals to reinforce multi-modal grounding and improve the navigation policy learning. 

\section{Error Analysis}
In this section, we further analyze the negative examples and showcase a few common errors in the vision-language navigation task. First, a common mistake comes from the misunderstanding of the natural language instruction. Figure~\ref{fig:neg_short} demonstrate such a qualitative example, where the agent successfully perceived the concepts ``hallway", ``turn left", and ``mirror" etc., but misinterpreted the meaning of the whole instruction. It turned left earlier and mistakenly entered the bathroom instead of the bedroom at Step 3. 

Secondly, failing to ground objects in the visual scene can usually result in an error. As shown in Figure~\ref{fig:neg_v} (a), the agent did not recognize the ``mannequins" in the end (Step 5) and stopped at a wrong place even though it executed the instruction pretty well. Similar in Figure~\ref{fig:neg_v} (b), the agent failed to detect the ``red ropes" at the beginning (Step 1) and thus took a wrong direction which also has the ``red carpet". Note that ``mannequins" is an out-of-vocabulary word in the training data; besides, both ``mannequins" and "red ropes" do not belong to the 1000 classes of the ImageNet~\cite{imagenet_cvpr09}, so the visual features extracted from a pretrain ImageNet model~\cite{he2016deep} are not able to represent them.  

In Figure~\ref{fig:neg_long}, we illustrate a long negative trajectory which our agent produced by following a relatively complicated instruction. In this case, the agent match ``the floor is in a circle pattern" with the visual scene, which seems to be another limitation of the current visual recognition systems. The above examples also suffer from the error accumulation issue as pointed out by Wang \etal~\cite{wang2018look}, where one bad decision leads to a series of bad decisions during the navigation process. Therefore, an agent capable of being aware of and recovering from errors is desired for future study.

\section{Trails and Errors}
Below are some trials and errors from our experimental experience, which are not the gold standard and used for reference only. 
\begin{itemize}
    \item We tried to incorporate dense bottom-up features as used in~\cite{anderson2018bottom}, but it hurt the performance on unseen environments. We think it is possibly because the navigation instructions require sparse visual representations rather than dense features. Dense features can easily lead to the overfitting problem. Probably more fine-grained detection results rather than dense visual features would help. 
    \item The performances are similar with or without positional encoding~\cite{vaswani2017attention} on the instructions.
    \item Pretrained ELMo embeddings~\cite{elmo} without fine-tuning hurts the performance. The summation of pretrained ELMo embeddings and task-specific embeddings has a similar effect of task-specific embeddings only. 
    \item It is not stable to only use the intrinsic reward to train the model. So we adopt the mixed reward for reinforcement learning, which works the best.
\end{itemize}

\end{document}